%% file: mainV2.tex
\documentclass[conference,a4paper]{IEEEtran}
\IEEEoverridecommandlockouts

\usepackage[hidelinks]{hyperref}
\usepackage[cmex10]{amsmath}
\usepackage{amssymb,amsfonts}
\usepackage{stfloats}

\usepackage[ruled,vlined]{algorithm2e}
\usepackage{graphicx}
\graphicspath{{figures/}}

\usepackage{booktabs}
\usepackage{siunitx}
\usepackage[numbers,compress]{natbib}
\usepackage{bm,bbm}
\usepackage{orcidlink}
\usepackage{xcolor}

\begin{document}

% --- TITLE ---
\title{\uppercase{DA-SegFormer: Damage-Aware Semantic Segmentation for Fine-Grained Disaster Assessment}}

% --- AUTHORS ---
\author{
    \IEEEauthorblockN{Kevin Zhu, William Tang, Raphael Hay Tene, Zesheng Liu, Nhut Le, and Maryam Rahnemoonfar}
    \IEEEauthorblockA{\textit{Bina Labs, Lehigh University}\\
    Bethlehem, PA, USA\\
    }
    \thanks{Copyright 2026 IEEE. Published in the 2026 IEEE International Geoscience and Remote Sensing Symposium (IGARSS 2026), scheduled for 9 - 14 August 2026 in Washington, D.C.. Personal use of this material is permitted. However, permission to reprint/republish this material for advertising or promotional purposes or for creating new collective works for resale or redistribution to servers or lists, or to reuse any copyrighted component of this work in other works, must be obtained from the IEEE. Contact: Manager, Copyrights and Permissions / IEEE Service Center / 445 Hoes Lane / P.O. Box 1331 / Piscataway, NJ 08855-1331, USA. Telephone: + Intl. 908-562-3966.}
\thanks{This version is the accepted manuscript submitted to arXiv. The final version will be published in the Proceedings of IGARSS 2026 and available via IEEE Xplore. For citation, please refer to the published version in IGARSS 2026.}
    % \thanks{\IEEEauthorrefmark{1} Correspondence to maryam@lehigh.edu} \IEEEauthorrefmark{1}
}

\maketitle

% --- ABSTRACT ---
\begin{abstract}
Rapid and accurate damage assessment following natural disasters is critical for effective emergency response. However, identifying fine-grained damage levels (e.g., distinguishing minor from major roof damage) in UAV imagery remains challenging due to the degradation of texture cues during resizing and extreme class imbalance. We propose DA-SegFormer, a damage-aware adaptation of the SegFormer architecture optimized for high-resolution disaster imagery. Our method introduces a Class-Aware Sampling strategy to guarantee exposure to rare damage features, and it integrates Online Hard Example Mining (OHEM) with Dice Loss to dynamically focus on underrepresented classes. In addition, we employ a resolution-preserving inference protocol that maintains native texture details. Evaluated on the RescueNet dataset, DA-SegFormer achieves 74.61\% mIoU, outperforming the baseline by 2.55\%. Notably, our improvements yield double-digit gains in critical damage classes: Minor Damage (+11.7\%) and Major Damage (+21.3\%).
\end{abstract}


% ===========================================================================
% SECTION 1: INTRODUCTION
% ===========================================================================
\section{Introduction}

Natural disasters such as hurricanes, floods, and earthquakes cause significant human and economic losses worldwide. In 2025 alone, the U.S. experienced 23 natural disasters costing approximately \$115 billion \cite{climatecentral2026}. A critical step in minimizing these losses is rapid and accurate damage assessment, which enables rescue teams to allocate resources efficiently, clear blocked transportation routes, and prioritize areas requiring immediate medical or structural attention. However, traditional manual assessment methods involving field supervision and damage reports are heavily time-consuming and often impossible to execute safely in severely affected areas \cite{rahnemoonfar2020}.

Semantic segmentation has emerged as a powerful tool for automated aerial damage assessment. Unlike image-level classification, semantic segmentation provides precise spatial localization of damage, enabling responders to identify exactly which structures are affected and to what degree. UAV imagery is particularly valuable, providing higher resolution than satellite imagery to capture detailed damage patterns like missing shingles, debris distribution, and partial structural collapse \cite{rahnemoonfar2020, safavi2021}.

The RescueNet dataset \cite{rahnemoonfar2020} provides a comprehensive benchmark for semantic segmentation in post-disaster scenarios. Collected after Hurricane Michael using UAVs at 200 ft AGL, the dataset contains approx. 2000 high-resolution images (3000 $\times$ 4000 pixels) with pixel-level annotations for 10 classes, capturing a wide variety of suburban damage profiles. Buildings are annotated at four distinct damage levels. Following \cite{rahnemoonfar2020}, these levels are explicitly defined as: No Damage (building unharmed), Minor Damage (parts damaged but coverable with blue tarp), Major Damage (significant structural damage requiring extensive repairs), and Total Destruction (complete failure of two or more major structural components).

Prior work on RescueNet \cite{rahnemoonfar2020} identified two fundamental challenges. First, differentiating between damage levels is extremely difficult from a top-down aerial view, as the visual differences between Minor and Major damage are often subtle texture variations. Standard computer vision pipelines frequently downsample inputs to satisfy memory constraints. In disaster assessment, this aggressive downsampling acts as a destructive low-pass filter, obliterating the very textures that define intermediate damage severities. Second, the dataset exhibits severe class imbalance: background pixels constitute 52.51\% of the data, while critical damage classes like Major Damage (1.68\%) and Total Destruction (1.44\%) are significantly underrepresented.

To address these domain-specific bottlenecks, we propose DA-SegFormer. Rather than a straightforward application of existing models, DA-SegFormer is a targeted architectural protocol designed to overcome the inherent failure modes of standard transformers when applied to fine-grained disaster imagery. Consequently, we introduce a synergistic approach: 1) A resolution-preserving inference protocol that maintains native textures, paired with a Class-Aware Sampling strategy during training to bridge the resolution gap. 2) A dynamic loss landscape integration utilizing Online Hard Example Mining (OHEM) and Dice Loss, strictly calibrated to force network attention onto the visually subtle decision boundaries between Minor and Major damage.

% ===========================================================================
% SECTION 2: RELATED WORK
% ===========================================================================
\section{Related Work}

\subsection{Semantic Segmentation for Disaster Assessment}

Deep learning has transformed automated damage assessment from aerial and satellite imagery. Rahnemoonfar et al. \cite{rahnemoonfar2020} introduced the RescueNet dataset and evaluated CNN-based architectures including PSPNet \cite{pspnet}, DeepLabv3+ \cite{deeplab}, and ENet \cite{enet} for comprehensive scene segmentation. Their experiments demonstrated that pyramid pooling approaches capture global context more effectively than standard encoder-decoder architectures for disaster imagery, achieving 79.43\% mIoU with PSPNet, as multi-scale pooling helps aggregate dispersed debris features. However, they also identified persistent challenges in distinguishing between intermediate damage levels (Minor vs. Major), where all models showed significantly lower performance.

Subsequent work explored attention mechanisms for this domain. Chowdhury and Rahnemoonfar \cite{chowdhury2021} applied self-attention methods to UAV imagery for damage assessment, demonstrating improved feature extraction for complex disaster scenes. Safavi et al. \cite{safavi2021} conducted a comparative study between real-time and non-real-time segmentation models on the FloodNet dataset \cite{floodnet}, showing that while lightweight models like UNet-MobileNetV3 achieve reasonable accuracy (59.3\% mIoU), non-real-time models like PSPNet (79.7\% mIoU) significantly outperform them.

Recent benchmarking on FloodNet \cite{rahnemoonfar2022} demonstrated that transformer-based architectures, specifically SegFormer, outperform CNN-based methods for post-disaster aerial imagery because their self-attention mechanisms inherently capture long-range dependencies across the large spatial dimensions of UAV images. This finding motivates our adoption of SegFormer as the backbone architecture. We also evaluate Mask2Former \cite{mask2former}, a universal segmentation transformer that uses masked attention to constrain cross-attention within predicted mask regions, as an additional strong baseline.

\subsection{Handling Class Imbalance}

Class imbalance is pervasive in remote sensing segmentation, where background and common classes dominate the pixel distribution. Standard Cross-Entropy loss causes models to converge toward predicting majority classes, ignoring rare but critical categories \cite{xu2023comparative}. 

Online Hard Example Mining (OHEM) \cite{ohem} addresses this by dynamically selecting high-loss examples during training. Rather than treating all pixels equally, OHEM focuses gradient updates on the most difficult samples, typically class boundaries and minority class pixels. Dice Loss \cite{diceloss}, derived from the Dice coefficient, provides complementary benefits by optimizing for region overlap on a per-class basis, giving equal weight to rare and common classes. Recent studies \cite{xu2023comparative} have shown that combining region-based losses with hard example mining improves minority class recall in segmentation tasks.

% ===========================================================================
% SECTION 3: METHODOLOGY
% ===========================================================================
\section{Methodology}

\subsection{SegFormer Architecture Details}

We adopt SegFormer \cite{segformer} as our backbone architecture. Unlike standard Vision Transformers (ViT) that generate single-resolution feature maps, SegFormer utilizes a hierarchical \textit{Mix Transformer} (MiT) encoder. This hierarchy is crucial for damage assessment as it captures both high-resolution coarse features (essential for locating small debris) and low-resolution fine features (essential for semantic context).

The MiT encoder generates multi-scale features through four stages, producing feature maps at resolutions $\{1/4, 1/8, 1/16, 1/32\}$ of the input image $H \times W$. A key innovation in MiT is the \textit{Overlapped Patch Merging} process. Standard ViT uses non-overlapping patch embeddings, which often results in a loss of local continuity around patch boundaries. SegFormer employs a convolution-based patch merging with an overlap, formalized as a convolution with kernel size $K=7$, stride $S=4$, and padding $P=3$. This preserves the local continuity of building edges and road networks, which are critical in the RescueNet dataset.

Furthermore, the encoder replaces fixed Positional Encodings (PE) with \textit{Mix-FFN}, a convolutional feed-forward network. This allows the model to handle variable input resolutions during inference without performance degradation, supporting our resolution-preserving inference strategy. The lightweight \textit{All-MLP} decoder aggregates these multi-scale features. Features $F_i$ from each stage $i$ are first passed through an MLP layer to unify the channel dimension, then upsampled to the $1/4$ resolution and concatenated as $F_{\text{concat}} = \text{Concat}(\forall i: \text{Upsample}(\text{MLP}(F_i)))$.

Finally, a prediction layer projects the concatenated features to the semantic segmentation mask $\in \mathbb{R}^{H/4 \times W/4 \times N_{\text{cls}}}$, where $N_{\text{cls}}$ represents the total number of semantic classes. We utilize the SegFormer-B4 variant, which features a deeper Stage 3 (27 transformer blocks) compared to lighter variants, providing the necessary capacity to model complex post-disaster scenes.

\subsection{Online Hard Example Mining (OHEM)}

To address the severe class imbalance in RescueNet (Table \ref{tab:class_stats}), we integrate Online Hard Example Mining into our training pipeline. OHEM dynamically identifies and prioritizes the most difficult pixels during each forward pass.

\begin{table}[h]
    \centering
    \caption{RescueNet Class Distribution. Damage classes constitute less than 8\% of total pixels combined.}\label{tab:class_stats}
    \begin{tabular}{l c | l c}
        \toprule
        \textbf{Class} & \textbf{Freq (\%)} & \textbf{Class} & \textbf{Freq (\%)} \\ \cmidrule(lr){1-2} \cmidrule(lr){3-4}
        Background & 52.51 & Major Damage & 1.68 \\
        Tree & 22.05 & Road-Blocked & 1.59 \\
        Water & 8.13 & Total Destruction & 1.44 \\
        Minor Damage & 2.63 & Pool & 0.06 \\ \bottomrule
    \end{tabular}
\end{table}

During each forward pass, we compute pixel-wise Cross-Entropy loss for all $N$ pixels in the batch. We then rank pixels by loss value and select the top $k=100,000$ hardest pixels to form the hard example set $\mathcal{K}$. This specific $k$ value represents approximately 10\% of the pixels in our 1024$\times$1024 training crops, ensuring the model receives a robust gradient signal from difficult regions without being overwhelmed by noisy outliers. The OHEM loss is computed only over this subset: $\mathcal{L}_{\text{OHEM}} = \frac{1}{k} \sum_{p \in \mathcal{K}} -\log(P(y_p | x_p))$, where $x_p$ represents the input pixel features, $y_p$ is the corresponding ground truth label, and $P(y_p | x_p)$ is the predicted probability of the true class. This approach forces the optimizer to focus on decision boundaries where the model is uncertain, such as the transitions between damage levels (e.g., Minor to Major) and edges between buildings and background. 

\subsection{Dice Loss}

While OHEM addresses pixel-level difficulty, Dice Loss handles class imbalance at the region level. Standard cross-entropy treats every pixel independently, which overwhelmingly biases the network toward the dominant background class. The Dice coefficient measures overlap between predicted and ground truth regions, defined as $\text{Dice}_c = (2 \sum_i p_{i,c} \cdot g_{i,c}) / (\sum_i p_{i,c} + \sum_i g_{i,c})$, where $p_{i,c}$ is the predicted probability for class $c$ at pixel $i$, and $g_{i,c}$ is the ground truth. Dice Loss is defined as $\mathcal{L}_{\text{Dice}} = 1 - \frac{1}{C} \sum_{c=1}^{C} \text{Dice}_c$, where $C$ is the total number of semantic classes. Importantly, Dice Loss computes overlap per class then averages across classes, giving equal weight to Minor Damage (2.63\% of pixels) and Background (52.51\%). This prevents the model from ignoring rare classes to minimize overall loss. Our total loss function combines both components: $\mathcal{L}_{\text{total}} = \mathcal{L}_{\text{Dice}} + \mathcal{L}_{\text{OHEM}}$.

\subsection{Class-Aware Sampling Strategy}

Standard random cropping is inefficient for disaster imagery as damage pixels constitute less than 8\% of the data. To counter this, we implement a Class-Aware Sampling strategy. Instead of uniform sampling, we enforce a bias where 50\% of training crops are centered on pixels belonging to underrepresented damage classes (Minor Damage, Major Damage, and Total Destruction). This guarantees that the model receives a consistent learning signal for critical features, ensuring that texture cues associated with damage are not overwhelmed by the dominant background classes.

\subsection{Resolution-Preserving Inference}

Prior work \cite{rahnemoonfar2022} showed fine-grained damage assessment depends on high-frequency texture cues (missing shingles, debris) destroyed by aggressive downsampling. Standard practice resizes entire images to fixed dimensions (e.g., 1024$\times$1024) during inference, creating a distribution shift when training uses higher resolution crops.

We address this through a resolution-preserving inference protocol. During inference, rather than resizing, we apply a sliding window with a 1024$\times$1024 patch size and stride $S=768$, resulting in a 25\% overlap. Predictions in overlapping regions are averaged uniformly to suppress boundary artifacts via $\mathbf{Y}_{\text{pred}} = \frac{1}{N_{ov}} \sum_{n=1}^{N_{ov}} f_\theta(x_n)$, where $N_{ov}$ is the number of overlapping predictions for a given pixel, and $f_\theta$ denotes our parameterized segmentation network. This ensures the model never encounters resolution-degraded inputs, maintaining consistency between the training and inference distributions.

% ===========================================================================
% SECTION 4: EXPERIMENT DETAILS
% ===========================================================================
\section{Experiment Details}

\subsection{RescueNet Dataset \& Evaluation Metrics}

We evaluate on the RescueNet dataset \cite{rahnemoonfar2020}, which contains 1,973 high-resolution UAV images (3000$\times$4000 pixels) captured after Hurricane Michael. Images are annotated with 10 semantic classes: Background, Water, Building-No-Damage, Building-Minor-Damage, Building-Major-Damage, Building-Total-Destruction, Road-Clear, Road-Blocked, Tree, and Vehicle. We use 1,591 images for training, 199 for validation, and 183 for testing. Preprocessing relies on the Class-Aware Sampling described in Section III-D to generate 1024$\times$1024 patches during training. 

To evaluate our proposed method, we utilize Intersection over Union (IoU) and Mean Intersection over Union (mIoU). Pixel accuracy is insufficient due to the severe class imbalance. The IoU for a specific class $c$ is defined as $\text{IoU}_c = \text{TP}_c / (\text{TP}_c + \text{FP}_c + \text{FN}_c)$, where $\text{TP}_c$, $\text{FP}_c$, and $\text{FN}_c$ represent the number of true positive, false positive, and false negative pixels respectively. The Mean IoU is calculated by averaging the IoU across all 11 classes: $\text{mIoU} = \frac{1}{11} \sum_{c=1}^{11} \text{IoU}_c$. We report pixel-level IoU scores for all classes to ensure transparency regarding the model's performance on underrepresented damage categories.

\subsection{Training Details}

We implement our method in PyTorch using the MMSegmentation framework. The SegFormer-B4 \cite{segformer} encoder is initialized with weights pretrained on ADE20K \cite{ade20k}. Training was conducted on NVIDIA L40S GPUs provided by Jetstream2 \cite{jetstream2} for 300 epochs with batch size 2. We use the AdamW optimizer \cite{adamw} with an initial learning rate of $6 \times 10^{-5}$ and a Cosine Annealing scheduler. Weight decay is set to 0.01. Data augmentation included random flips, rotations, and photometric distortions to improve model generalization. For OHEM, we select the top $k=100,000$ hardest pixels per batch. The Dice Loss and OHEM Loss are combined with equal weighting. For comparison, we train a baseline SegFormer-B4 \cite{segformer} with standard Cross-Entropy loss and input resizing to 1024$\times$1024.

% ===========================================================================
% SECTION 5: RESULTS
% ===========================================================================
\section{Results}

% --- DECLARE TABLE 2 FIRST TO FORCE IT TO THE TOP ---
\begin{table*}[t]
    \centering
    \caption{Per-Class IoU Comparison. DA-SegFormer yields critical gains in Minor (+11.7\%) and Major (+21.3\%) Damage compared to baselines.}\label{tab:results}
    \vspace{1.5mm}
    \small
    \setlength{\tabcolsep}{4.5pt}
    \begin{tabular}{l c c c c c c c c c c c c}
        \toprule
        & & & \multicolumn{4}{c}{\textbf{Building Damage}} & \multicolumn{2}{c}{\textbf{Road}} & & & & \\
        \cmidrule(lr){4-7} \cmidrule(lr){8-9}
        \textbf{Method} & \textbf{Background} & \textbf{Water} & \textbf{None} & \textbf{Minor} & \textbf{Major} & \textbf{Destr.} & \textbf{Clear} & \textbf{Blocked} & \textbf{Tree} & \textbf{Pool} & \textbf{Vehicle} & \textbf{mIoU} \\ 
        \midrule
        Mask2Former \cite{mask2former} & \textbf{85.77} & 88.23 & \textbf{72.89} & 60.23 & 61.20 & 60.85 & 71.21 & \textbf{78.37} & 45.78 & 80.27 & \textbf{81.74} & 71.50 \\
        SegFormer-B4 \cite{segformer} & 85.70 & \textbf{89.60} & 71.20 & 61.20 & 50.80 & 60.20 & \textbf{83.20} & 45.80 & \textbf{82.80} & 84.90 & 77.20 & 72.06 \\
        DA-SegFormer (ours) & 85.02 & 88.80 & 69.06 & \textbf{72.93} & \textbf{72.05} & \textbf{63.64} & 82.88 & 41.28 & 81.25 & \textbf{87.53} & 76.22 & \textbf{74.61} \\
        \midrule
        Gain (vs Segformer-B4) & -0.68 & -0.80 & -2.14 & \textbf{+11.73} & \textbf{+21.25} & \textbf{+3.44} & -0.32 & -4.52 & -1.55 & \textbf{+2.63} & -0.98 & \textbf{+2.55} \\
        \bottomrule
    \end{tabular}
\end{table*}

% --- DECLARE FIGURE 1 SECOND SO IT SITS UNDER THE TABLE ---
\begin{figure*}[t]
    \centering
    % Ensure image is not too wide
    % \hspace*{-3cm} 
    \includegraphics[width=0.8\textwidth]{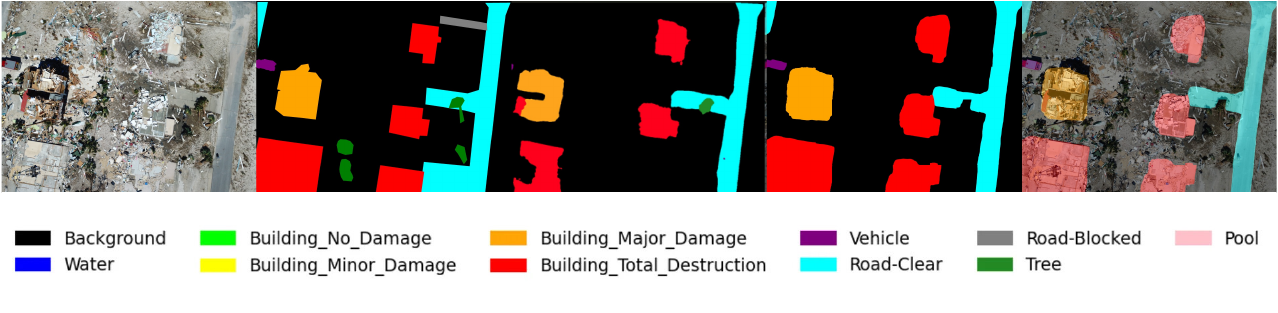}
    % \vspace{1mm}
    % \includegraphics[width=0.85\textwidth]{legend.jpg}
    \caption{Qualitative comparison on RescueNet test images. Columns: (a) Original Image, (b) Ground Truth, (c) SegFormer Baseline, (d) DA-SegFormer, (e) Overlay. Our method produces sharper building boundaries and more accurate damage level classification.}
    \label{fig:qualitative}
\end{figure*}

We compare our improved SegFormer against the baseline SegFormer-B4 \cite{segformer} trained with standard Cross-Entropy loss and 1024$\times$1024 resizing. All models use identical encoder initialization and training epochs.

\subsection{Quantitative Results}

Table \ref{tab:results} presents per-class IoU and mean IoU. DA-SegFormer achieves 74.61\% mIoU, a 2.55\% improvement over the baseline SegFormer (72.06\%). The most significant improvements occur in the damage severity classes critical for emergency response: Minor Damage improves by 11.73\% (61.20\% to 72.93\%) and Major Damage by 21.25\% (50.80\% to 72.05\%). These classes require distinguishing subtle texture differences visible only at high resolution; this is precisely what our resolution-preserving inference and hard example mining address. 

Mask2Former \cite{mask2former} achieves 71.50\% mIoU---falling 0.56\% short of the SegFormer baseline and 3.11\% short of DA-SegFormer. Notably, while Mask2Former performs adequately on general background features, it struggles to cleanly delineate the subtle boundaries of structural damage. It achieves only 60.23\% on Minor Damage and 61.20\% on Major Damage, falling significantly short of DA-SegFormer's 72.93\% and 72.05\% in those respective categories. While legacy CNNs achieve strong aggregate scores by merging intermediate damage levels \cite{rahnemoonfar2020}, standard architectures consistently underperform on our strict 11-class fine-grained formulation. This highlights the necessity of our targeted architectural protocol.

\subsection{Computational Efficiency}

Table \ref{tab:comp_cost} details the number of parameters, Floating Point Operations (GFLOPs), and inference speed measured in Frames Per Second (FPS) on a single NVIDIA L40S GPU.

% --- DECLARE TABLE 3 HERE WITH HTBP ---
\input{table_3}

DA-SegFormer requires higher computational resources (684.31 GFLOPs) than standard resizing approaches due to the resolution-preserving 1024$\times$1024 sliding-window inference protocol. While its base architecture is lighter than universal models like Mask2Former, DA-SegFormer's frames-per-second is bounded at 3.00 because our protocol requires processing approximately 20 overlapping crops per single UAV image. This throughput remains viable for offline post-disaster analysis where batch processing is standard. Extreme real-time constraints are often secondary to assessment accuracy. The increased inference time is a necessary trade-off to preserve the high-frequency texture cues required to distinguish accurately between Minor and Major damage levels.

\subsection{Qualitative Results}

As shown in Figure \ref{fig:qualitative}, baseline SegFormer frequently confuses Minor and Major damage, producing fragmented predictions within single uniform roofs. In contrast, DA-SegFormer produces coherent, building-level predictions with sharper boundaries, successfully distinguishing intact roof segments from severe structural failure. The baseline also misses localized debris due to downsampling texture degradation. Conversely, our resolution-preserving inference maintains the high-frequency features necessary for accurate contiguous segmentation across complex post-disaster scenes.

% ===========================================================================
% SECTION 6: CONCLUSION
% ===========================================================================
\section{Conclusion}

We presented DA-SegFormer, a damage-aware adaptation of the SegFormer architecture for fine-grained disaster assessment. By integrating Online Hard Example Mining with Dice Loss, we address the severe class imbalance inherent to disaster imagery. Our training-aligned inference preserves native textures critical for distinguishing subtle damage. On RescueNet, DA-SegFormer achieves 74.61\% mIoU, outperforming the baseline by 2.55\%. Notably, it yields double-digit improvements on critical classes: Minor (+11.7\%) and Major (+21.3\%) damage \cite{rahnemoonfar2020}. These results confirm algorithmic handling of imbalance and texture preservation are as crucial as network architecture. Future work will explore multi-temporal damage assessment.

% ===========================================================================
% ACKNOWLEDGMENT
% ===========================================================================
\section*{Acknowledgment}
Supported by NSF (\#2423211, \#2401942), the Consortium for Enhancing Resilience and Catastrophe Modeling, and ACCESS. Jetstream2 resources at Indiana Univ. via CIS251039 (NSF \#2138259, \#2138286, \#2138307, \#2137603, \#2138296).

% ===========================================================================
% BIBLIOGRAPHY
% ===========================================================================
% \clearpage

\end{document}

%% file: table_3.tex
\begin{table}[h]
    \centering
    \caption{Computational Cost Comparison (Native inference resolutions).}\label{tab:comp_cost}
    \begin{tabular}{l c c c}
        \toprule
        \textbf{Method} & \textbf{Params (M)} & \textbf{GFLOPs} & \textbf{FPS} \\ 
        \midrule
        SegFormer-B4 \cite{segformer} & \textbf{64.00} & \textbf{478.85} & \textbf{30.53}  \\
        Mask2Former \cite{mask2former} & 215.45 & 1943.92 & 6.65 \\
        DA-SegFormer-B4 (ours) & \textbf{64.00} & 684.31 & 3.00 \\
        \bottomrule
    \end{tabular}
\end{table}

%% file: mainV2.bbl
\begin{thebibliography}{99}

\bibitem{climatecentral2026}
Climate Central, ``2025 in Review: U.S. Billion-Dollar Disasters,'' \textit{Climate Central}, Jan. 8, 2026. [Online]. Available: https://www.climatecentral.org/climate-matters/2025-in-review

\bibitem{rahnemoonfar2020}
M. Rahnemoonfar, T. Chowdhury, R. Murphy, and O. Fernandes, 
``Comprehensive semantic segmentation on high resolution UAV 
imagery for natural disaster damage assessment,'' in \textit{Proc. 
IEEE Int. Conf. Big Data}, Atlanta, GA, USA, 2020, pp. 3726--3735.

\bibitem{floodnet}
M. Rahnemoonfar, T. Chowdhury, A. Sarkar, M. Varshney, M. Yari, and R. Murphy,
``FloodNet: A high resolution aerial imagery dataset for post-flood 
scene understanding,'' \textit{IEEE Access}, vol. 9, pp. 89644--89654, 2021.

\bibitem{segformer}
E. Xie, W. Wang, Z. Yu, A. Anandkumar, J. M. Alvarez, and P. Luo,
``SegFormer: Simple and efficient design for semantic segmentation 
with transformers,'' in \textit{Proc. NeurIPS}, 2021.

\bibitem{chowdhury2021}
T. Chowdhury and M. Rahnemoonfar, ``Attention based semantic 
segmentation on UAV dataset for natural disaster damage 
assessment,'' in \textit{Proc. IEEE Int. Geosci. Remote Sens. 
Symp. (IGARSS)}, Brussels, Belgium, 2021, pp. 2325--2328.

\bibitem{safavi2021}
F. Safavi, T. Chowdhury, and M. Rahnemoonfar, ``Comparative 
study between real-time and non-real-time segmentation models 
on flooding events,'' in \textit{Proc. IEEE Int. Conf. Big Data}, 
Orlando, FL, USA, 2021, pp. 4199--4207.

\bibitem{rahnemoonfar2022}
M. Rahnemoonfar \textit{et al.}, ``Real-time semantic segmentation 
of aerial imagery for emergency response,'' \textit{IEEE J. Sel. 
Topics Appl. Earth Observ. Remote Sens.}, vol. 16, pp. 4--20, 2022.

\bibitem{ohem}
A. Shrivastava, A. Gupta, and R. Girshick, ``Training region-based 
object detectors with online hard example mining,'' in \textit{Proc. 
IEEE CVPR}, 2016, pp. 761--769.

\bibitem{diceloss}
F. Milletari, N. Navab, and S.-A. Ahmadi, ``V-Net: Fully convolutional 
neural networks for volumetric medical image segmentation,'' in 
\textit{Proc. Int. Conf. 3D Vis. (3DV)}, 2016, pp. 565--571.

\bibitem{xu2023comparative}
Z. Xu \textit{et al.}, ``A comparative study of loss functions 
for road segmentation in remote sensing imagery,'' \textit{Int. 
J. Appl. Earth Obs. Geoinf.}, vol. 116, 2023.

\bibitem{pspnet}
H. Zhao, J. Shi, X. Qi, X. Wang, and J. Jia, ``Pyramid scene 
parsing network,'' in \textit{Proc. IEEE CVPR}, 2017, pp. 2881--2890.

\bibitem{deeplab}
L.-C. Chen, Y. Zhu, G. Papandreou, F. Schroff, and H. Adam, 
``Encoder-decoder with atrous separable convolution for semantic 
image segmentation,'' in \textit{Proc. ECCV}, 2018, pp. 801--818.

\bibitem{enet}
A. Paszke, A. Chaurasia, S. Kim, and E. Culurciello, ``ENet: A 
deep neural network architecture for real-time semantic 
segmentation,'' \textit{arXiv preprint arXiv:1606.02147}, 2016.

\bibitem{adamw}
I. Loshchilov and F. Hutter, ``Decoupled weight decay 
regularization,'' in \textit{Proc. ICLR}, 2019.

\bibitem{ade20k}
B. Zhou, H. Zhao, X. Puig, S. Fidler, A. Barriuso, and A. Torralba,
``Scene parsing through ADE20K dataset,'' in \textit{Proc. IEEE 
CVPR}, 2017, pp. 633--641.

\bibitem{jetstream2}
D. Hancock \textit{et al.}, ``Jetstream2: Accelerating cloud 
computing via Jetstream,'' in \textit{Proc. PEARC}, 2021.

\bibitem{mask2former}
B. Cheng, I. Misra, A. G. Schwing, A. Kirillov, and R. Garg,
``Masked-attention mask transformer for universal image segmentation,''
in \textit{Proc. IEEE CVPR}, 2022, pp. 1290--1299.

\end{thebibliography}
